\documentclass[conference]{IEEEtran}
\IEEEoverridecommandlockouts
\usepackage[margin=1in]{geometry}
\usepackage{cite}
\usepackage{amsmath,amssymb,amsfonts}
\usepackage{graphicx}
\usepackage{textcomp}
\usepackage{xcolor}
\usepackage{hyperref}
\usepackage{algorithm}
\usepackage{subcaption}
\usepackage{graphicx}
\usepackage{float}
\usepackage{caption}
\usepackage{paralist}
\usepackage{tcolorbox}
\usepackage{mdframed}
\usepackage{listings}
\usepackage{multirow}
\usepackage{enumitem}
\usepackage{tikz}
\usetikzlibrary{arrows.meta,positioning,shapes.geometric,matrix,calc}

\usepackage{pgfplots}
\pgfplotsset{compat=1.18}

\usepackage{titlesec}

\titlespacing*{\section}
{0pt}{4pt}{4pt}

\titlespacing*{\subsection}
{0pt}{2pt}{2pt}

\newtheorem{problem}{Problem}

\mdfsetup{
  linewidth=0.5pt,
  innertopmargin=2pt,
  innerbottommargin=2pt,
  innerleftmargin=4pt,
  innerrightmargin=4pt,
  backgroundcolor=white,
  skipabove=0pt,
  skipbelow=0pt
}
\usepackage[noend]{algpseudocode}
\begin{document}

\title{\textbf{Real-Time Optical Communication Using Event-Based Vision with Moving Transmitters}}

\author{Harmeet Dhillon, Pranay Katyal, Brendan Long, Rohan Walia, Matthew Cleaveland, Kevin Leahy
\thanks{DISTRIBUTION STATEMENT A. Approved for public release. Distribution is unlimited. This material is based upon work supported by the Under Secretary of War for Research and Engineering under Air Force Contract No. FA8702-15-D-0001 or FA8702-25-D-B002. Any opinions, findings, conclusions or recommendations expressed in this material are those of the author(s) and do not necessarily reflect the views of the Under Secretary of War for Research and Engineering. © 2026 Massachusetts Institute of Technology. Delivered to the U.S. Government with Unlimited Rights, as defined in DFARS Part 252.227-7013 or 7014 (Feb 2014). Notwithstanding any copyright notice, U.S. Government rights in this work are defined by DFARS 252.227-7013 or DFARS 252.227-7014 as detailed above. Use of this work other than as specifically authorized by the U.S. Government may violate any copyrights that exist in this work.}
\thanks{H. Dhillon, P. Katyal, R. Walia and K. Leahy are with the Robotics Engineering Department, Worcester Polytechnic Institute, Worcester, MA, USA.
        {\tt\small \{hsdhillon,pkatyal,rwalia,kleahy\} @wpi.edu}}%
\thanks{B. Long and M. Cleaveland are with MIT Lincoln Laboratory, Lexington, MA, USA.
        {\tt\small \{brendan.long,matthew.cleaveland\} @ll.mit.edu}}%
}

\maketitle

\begin{abstract}
In multi-robot systems, traditional radio frequency (RF) communication struggles with contention and jamming. Optical communication offers a strong alternative.
However, conventional frame-based cameras suffer from limited frame rates, motion blur, and reduced robustness under high dynamic range lighting. Event cameras support microsecond temporal resolution and high dynamic range, making them extremely sensitive to scene changes under fast relative motion with an optical transmitter. Leveraging these strengths, we develop a complete optical communication system capable of tracking moving transmitters and decoding messages in real time. Our system achieves over 95\% decoding accuracy for text transmission during motion by implementing a Geometry-Aware Unscented Kalman Filter (GA-UKF), achieving 7× faster processing speed compared to the previous state-of-the-art method, while maintaining equivalent tracking accuracy at transmitting frequencies $\geq$ 1 kHz.
\end{abstract}

\section{Introduction}

Multi-robot systems and swarm robotics hold tremendous promise in a variety of domains, such as search and rescue~\cite{drew2021multi}, underwater exploration~\cite{zhou2022}, infrastructure inspection~\cite{halder2023robots}, and agriculture~\cite{dutta2021}, among others. Communication and information sharing is critical to the success of multi-robot systems. The primary modality used for communication in multi-robot systems is radio frequency (RF) technology. RF offers ease of omnidirectional transmission through the air. However, RF leads to contention, which requires deployment of dedicated protocols to manage bandwidth~\cite{abramson1970aloha,degesys2007desync}. Additionally, adversarial interference (jamming) poses security concerns for multi-robot systems using RF technology ~\cite{pirayesh2022jamming}.

Optical communication has the potential to serve as a strong alternative to RF based communication. It eliminates the need for broadcasting signals, and can be applied to a variety of applications spanning from localization to swarm formation and control \cite{underwater_optical_comm_application}. However, optical communication also has limitations. For example, lasers enable high-bandwidth communication between robots but require precise localization of the transmitter and receiver~\cite{tsujimura2018spacial}, making this method unsuitable for dynamic or ad-hoc communication among robots. Similarly, LED modulation has been successfully used for robot communication~\cite{wang2022smart}, but only with a single stationary transmitter-receiver pair. 

Frame-based camera systems perform well in passive communication settings, such as feature matching across successive frames to track a static optical transmitter. However, active communication tasks, such as decoding high-frequency modulation or tracking a moving transmitter, require sensing changes in near-continuous time. Due to frame acquisition latency and limited temporal resolution, conventional frame-based cameras struggle in these scenarios \cite{event_vision_survey}. In contrast, event cameras asynchronously report pixel-level brightness changes, providing microsecond temporal resolution and high dynamic range \cite{gallego_event_survey}. This makes them better suited for active communication tasks. 


Recently, Wang et al. demonstrated robust decoding of signals from an optical transmitter (LED) using event cameras\cite{wang2022smart}. However, achieving highly accurate real-time optical communication under relative motion between the transmitter and receiver remains an open problem. Addressing this challenge is crucial for enabling practical optical communication between moving robots.





\begin{figure}
    \centering
    \begin{tikzpicture}

      \node[anchor=south west, inner sep=0] (main) at (0,0)
        {\includegraphics[width=\linewidth, height=4.75cm]{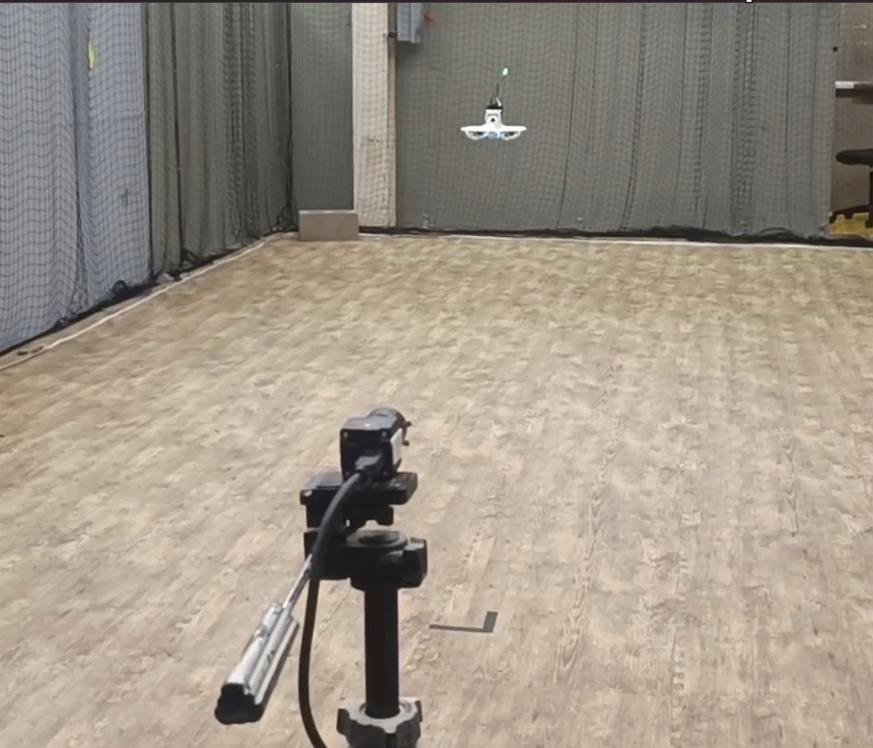}};
      \begin{scope}[x={(main.south east)}, y={(main.north west)}] 
    
        \coordinate (roiNE) at (0.61,0.95);
        \coordinate (roiSW) at (0.52,0.75);
    
        \draw[red, very thick] (roiSW) rectangle (roiNE);
    
        \node[anchor=north east, inner sep=0] (inset) at (0.98,0.75)
          {\includegraphics[width=0.35\linewidth]{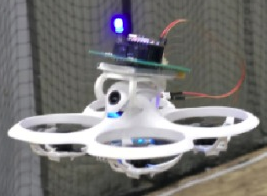}};
    
        \draw[black, thick] (inset.south west) rectangle (inset.north east);
    
        \draw[red, very thick] (inset.south west) rectangle (inset.north east);
    
        \draw[red, thick] (roiNE) -- (inset.north east);
        \draw[red, thick] (roiSW) -- (inset.south west);
    
      \end{scope}
    \end{tikzpicture}
    \caption{Hardware experimental setup. Bottom: Prophesee EVK4 event camera capturing a strobing LED mounted on the drone. Inset: Close-up view of the drone with the LED.}
    \label{fig:hardware}
\end{figure}
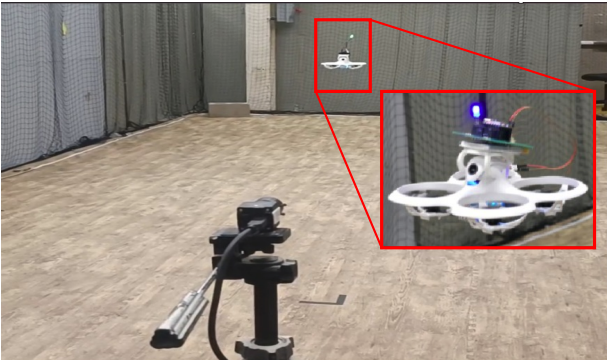

\subsection{Related Work}
Prior work has demonstrated promising results in decoding optical signals. In ~\cite{wang2022smart}, Wang et al. report high-fidelity, real-time decoding of LED-based optical transmissions using event cameras. This approach is effective when both the transmitter (LED) and receiver (event camera) are stationary. However, communication deteriorates sharply when the relative motion between transmitter-receiver starts to become non-zero.

Lossless decoding under relative motion requires accurately tracking the transmitter. Due to optical defocus, motion blur, and intrinsic sensor characteristics, a single optical transmitter, such as a LED, typically spans multiple pixels when projected onto the image plane. When observed using an event camera, it produces a spatial cluster of events rather than a single pixel activation. This cluster, called a \emph{blob}, is well-approximated by an elliptical structure in the image plane\cite{async_blob_wang}. Blob tracking has been implemented in~\cite{async_blob_wang} using an Extended Kalman Filter (EKF). However, as discussed in the next section, the computational complexity of this method scales linearly with the event rate. This results in a bottleneck at high frequencies. 



In this work, we address the limitations of \cite{wang2022smart} and \cite{async_blob_wang} to develop a practical optical transmission system that can be deployed on mobile robots in real-time. Our key contributions are:

\begin{itemize}
\item We introduce the Geometry-Aware UKF (GA-UKF), which exploits both the geometric structure of the blob (its coordinates, shape, and orientation) and temporal dynamics of the events to accurately estimate the transmitter (LED) position under motion in real time.

\item Using GA-UKF, we build an optical communication system capable of asynchronous transmitter tracking, spatial filtering, temporal transformation estimation, and batch processing to enable robust optical communication between a moving optical transmitter-receiver pair.

\item We achieve tracking speeds of up to 10{,}000 pixels/s for a moving optical transmitter, with a $7\times$ reduction in processing time compared to the baseline method in~\cite{async_blob_wang}.

\item We demonstrate over 90\% decoding accuracy for optically transmitting a stream of text under relative transmitter–receiver motion.
\end{itemize}
\section{Problem Formulation}
In this section, we highlight the limitations of the current state-of-the-art method for state estimation of optical transmitters, and how explicit modeling of geometry of the estimated state is essential to obtain accurate estimates under motion.

\subsection{Transmitter tracking using blob model}

Wang et al. use an Extended Kalman Filter (EKF) to sequentially process event camera data for tracking an LED ~\cite{async_blob_wang}. To capture the uncertainty about it's position in the 2D image plane, they model the optical transmitter as an ellipsoid (\textit{blob}):
\begin{equation}
    \mathbf{x} = [x, y, v_x, v_y, \theta, q, \lambda_1, \lambda_2]^\top,
    \label{eq:async_state_vector}
\end{equation}
where $(x,y)$ denotes the blob centroid, $(v_x, v_y)$ represent translational velocities, $\theta$ denotes orientation, $q$ is the angular velocity, and ($\lambda_1$, $\lambda_2$) correspond to the semi-axes lengths of the blob's ellipsoid.

In this work, the EKF incrementally updates the blob state belief as events arrive. Consequently, processing latency grows with the number of events in each event data packet. This is viable for passive tracking, where event data packets have low number of events. However, performance may degrade in active tracking scenarios if the computational cost of sequentially processing exceeds the inter-packet transmission interval. On the other hand, increasing the batch size of events per update introduces stronger nonlinearities, thereby degrading the linearization approximation accuracy. 

For real-time active blob tracking, an estimator needs to process asynchronous, noisy stream of events with minimum latency. This can be achieved by processing an entire data packet at once instead of processing each event individually. However, the ellipsoidal shape of the blob is prone to distortion if the entire spatial distribution is accumulated for processing at once. Therefore, relying on spatial information alone is not enough.

By incorporating temporal information and assigning higher weights to more recent events, the estimated distribution better reflects the true instantaneous geometry of the blob. Under such temporally adaptive weighting, the event cluster more closely resembles an ellipse. Therefore,  the maximum likelihood of the blob can be represented using the following multivariate Gaussian model \cite{async_blob_wang}~\cite{johnson2007applied}:
\begin{equation}
    \begin{aligned}
        f(\mathbf{x}(p,t)) &=
            \frac{1}{(2\pi)^{n/2}|\Sigma(p,t)|^{1/2}}
            \, e^{-\tfrac12
                \boldsymbol{\delta}^{T}
                \Sigma(p,t)^{-1}
                \boldsymbol{\delta}}, \\[6pt]
        \boldsymbol{\delta} &=
            \mathbf{x}(p,t)-\boldsymbol{\mu}(p,t).
    \end{aligned}
    \label{blob_gaussian}
\end{equation}
Here, $\mathbf{x}(p,t) \in \mathbb{R}^2$ denotes the spatial location of an event within the blob region, $\boldsymbol{\mu}(p,t) \in \mathbb{R}^2$ represents the estimated center of the transmitter (LED), $p \in \{+1,-1\}$ denotes event polarity, and $t$ denotes time. The matrix $\Sigma(p,t) \in \mathbb{S}_{++}^{2}$ is the symmetric positive definite covariance matrix of the blob distribution where $\mathbb{S}_{++}^{2}$ represents a $2\times2$ manifold of symmetric positive definite matrices, which characterizes the shape and orientation of the Gaussian distribution and therefore defines the geometric structure of the blob~\cite{johnson2007applied}. Henceforth, any reference to the shape or characteristics of the covariance matrix $\Sigma$ equivalently refers to the geometry of the blob, and vice-versa.

When the only available measurements consist of a asynchronous stream of events generated by the moving transmitter, it makes the state-estimation problem more challenging. Events get corrupted by motion-induced spatial spread and background noise~\cite{gallego_event_survey}. This results in a highly irregular and noisy observation process. There is no access to dense intensity images or structured pixel information. Instead, all inference must rely exclusively on the spatio-temporal distribution of events surrounding the transmitter. Since the motion of the object is independent of the camera’s control input, an accurate dynamic model is not available, which introduces additional nonlinear effects. As the EKF relies on first-order linearization, its state-estimation accuracy degrades as the effective degree of nonlinearity increases\cite{julier2004unscented}.

\subsection{Geometry-Aware State Estimation}

\begin{figure}[t]
  \centering
  \begin{subfigure}[t]{0.525\columnwidth}
    \centering
    \includegraphics[width=\linewidth]{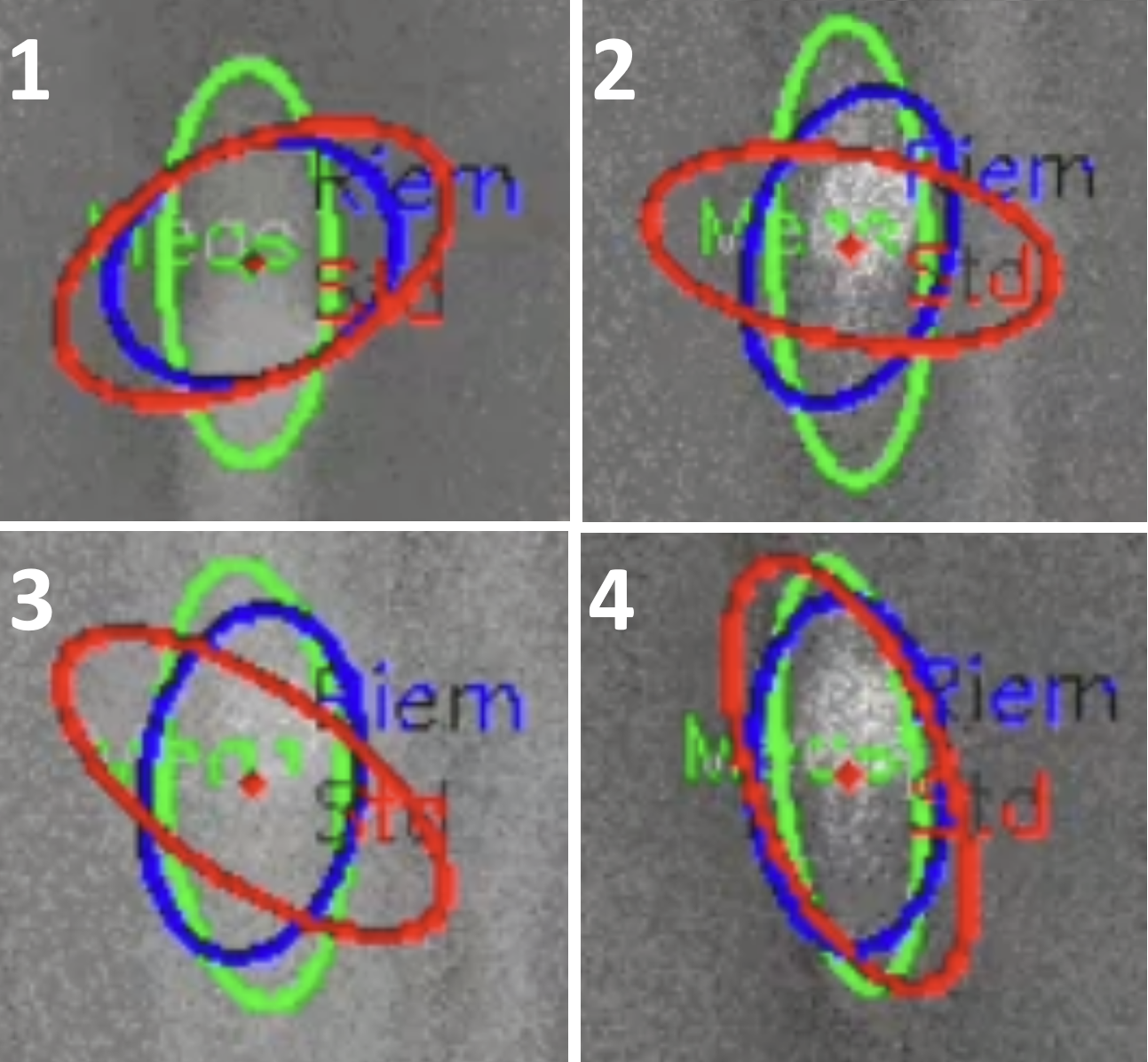}
  \end{subfigure}%
  \hspace{0.01\columnwidth}%
  \begin{subfigure}[t]{0.42\columnwidth}
  \centering
        \begin{tikzpicture}[xscale=0.5, yscale=0.97]

          \def\ellA{0.8}     
          \def\ellB{0.3}     
        
          \def\arrowR{1.0}    

          \def\headingY{1.2} 
        
          \draw[black] (-3.5, -2.0) rectangle (3.5, 2.0);
    
          \draw[black!50] (3.5, 2.0) -- (-3.5, -2.0);
    
          \begin{scope}
            \clip (-3.5, 2.0) -- (3.5, 2.0) -- (-3.5, -2.0) -- cycle;
    
            \begin{scope}[xshift=-1.5cm, yshift=0.6cm]
    
              \draw[green, thick, rotate=90] (0,0) ellipse (\ellA cm and \ellB cm);
              \draw[red, thick, dashed, rotate=35] (0,0) ellipse (\ellA cm and \ellB cm);
              \draw[red, thick, dashed, rotate=145] (0,0) ellipse (\ellA cm and \ellB cm);
    
              \draw[-{Stealth[length=6pt]}, red, thick] ([xshift=0.25cm]22.5:\arrowR) arc[start angle=22.5, end angle=-22.5, radius=\arrowR];
              \draw[-{Stealth[length=6pt]}, red, thick] ([xshift=-0.25cm]-157.5:\arrowR) arc[start angle=-157.5, end angle=-202.5, radius=\arrowR];
    
              \node[red, font=\small] at (1.4, 0.7) {initial};
              \node[font=\small\bfseries, align=center] at (0.5, \headingY) {Linear Mapping};
    
            \end{scope}
          \end{scope}
    
          \begin{scope}
            \clip (-3.5, -2.0) -- (3.5, -2.0) -- (3.5, 2.0) -- cycle;
    
            \begin{scope}[xshift=1.0cm, yshift=-0.6cm]

              \draw[green, thick, rotate=90] (0,0) ellipse (\ellA cm and \ellB cm);
              \draw[blue, thick, dashed, rotate=50] (0,0) ellipse (\ellA cm and \ellB cm);
    
              \draw[-{Stealth[length=6pt]}, blue, thick] (180:\arrowR) arc[start angle=180, end angle=247.5, radius=\arrowR];
    
              \node[blue, font=\small] at (1.5, 0.75) {initial};
              \node[font=\small\bfseries, align=center] at (0.5, -\headingY) {$\mathbb{S}^1$ Wrapped};
    
            \end{scope}
          \end{scope}
    
        \end{tikzpicture}%
    \end{subfigure}

  \caption{\small
Left: Estimated blob orientation using linear (Euclidean) mapping (red) versus an $\mathbb{S}^1$-wrapped update (blue), with the ground-truth orientation shown in green. 
Right: Illustration of the corresponding rotation trajectories. Dashed curves denote intermediate states and solid ellipses denote final estimates. The Euclidean update follows a longer rotation path, whereas the $\mathbb{S}^1$-wrapped update follows the shorter, geometrically consistent path. Dynamic visualization is presented in the supplementary video.}
  \label{fig:linear_rotation}
\end{figure}
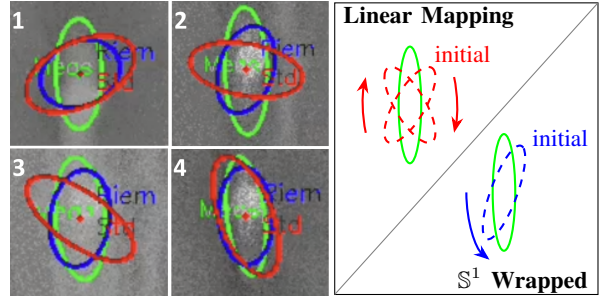

Linearization errors can be avoided by using an Unscented Kalman Filter (UKF), which relies on sigma-point propagation to approximate the belief distribution of the blob's state  \cite{julier2004unscented}. However, directly applying a UKF in Euclidean space defies the actual geometrical structure of the state, particularly the shape and orientation of the blob. Firstly, naive UKF state estimates could result in negative sigma-point values associated with axis-lengths of the blob, which is physically meaningless as $\lambda_1, \lambda_2 \in \mathbb{R_+}$. This can happen when sigma-point dispersion around a small positive mean is large, or when asymmetrically distributed and weighted sigma points bias the estimate away from the positive definite region. Secondly, the shortest arc between two blob orientations exists on the $\mathbb{S}^1$ manifold. Therefore, processing orientation in Euclidean space may lead to incorrect estimation of orientation trajectories, as shown in Fig.~\ref{fig:linear_rotation}.

These errors arise due to neglecting the intrinsic geometry of the covariance matrix in \eqref{blob_gaussian}. Since the covariance matrix $\Sigma$ is symmetric positive definite (SPD), it resides on a smooth Riemannian manifold $\mathbb{S}_{++}^2$ endowed with an affine-invariant metric ~\cite{bhatia2007positive}, rather than in a flat Euclidean space. This geometric structure necessitates manifold-aware filtering techniques to ensure accurate state updates.

In this work, we adopt event-vision based decoding proposed in ~\cite{wang2022smart} to demonstrate real-time optical communication. The performance of this method is directly tied to the accuracy of the estimated transmitter position. This implies minimizing uncertainty about the transmitter’s state, i.e. the blob’s geometric spread. As discussed above, such estimates demand a state estimation framework capable of handling high-frequency asynchronous inputs while preserving geometric consistency and accounting for nonlinearities. Essentially, such a framework must solve Problem \ref{pb:main}:

\begin{problem}\label{pb:main}
Given a blob state of the form \eqref{eq:async_state_vector}, estimate
${\mathbf{x}}$ such that ${\lambda}_1 \ge 0$ and ${\lambda}_2 \ge 0$, and the orientation
${\theta}$ remains on $\mathbb{S}^1$ (i.e., does not wrap across $2\pi$ / $360^\circ$).
\end{problem}





\section{Geometry-Aware Unscented Kalman Filter} \label{sec:methodology}
In order to solve Problem \ref{pb:main}, we propose the Geometric-Aware Unscented Kalman Filter (GA-UKF) algorithm. GA-UKF explicitly accounts for the underlying manifold geometry of the blob state. Additionally, we develop a spatial filter which filters all the events that are not in the region of interest. This reduces memory usage and latency (if input data is processed multiple times) in the process.

\begin{figure*}[t]
    \centering
    \includegraphics[width=1\textwidth]{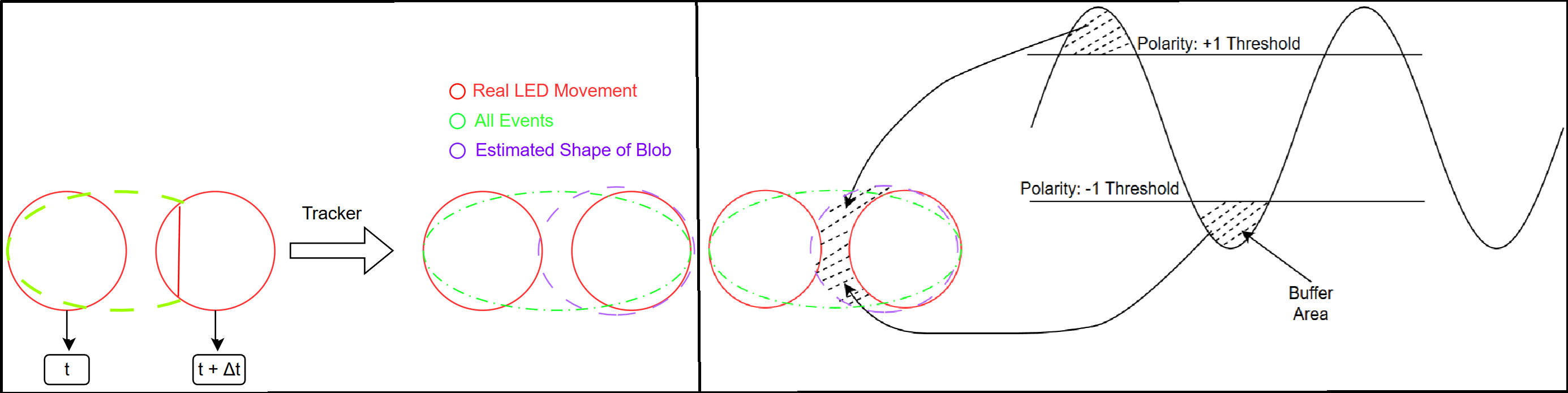}
    \caption{\small Left: GA-UKF tracks LED position and shape amid motion. The green ellipse represents all events in one batch update, the red circle represents true LED position, and the purple ellipse shows the estimated blob. Right: Continuous-time signal reconstruction using dual-threshold hysteresis, robustly recovering the binary sequence despite motion and noise.}
    \label{fig:finalIllus}
\end{figure*}

\begin{figure}[t]
\centering
\resizebox{\columnwidth}{!}{%
\begin{tikzpicture}[
    font=\small,
    >=Latex,
    node distance=10mm and 10mm,
    box/.style={draw, rounded corners=2pt,
                minimum width=18mm, minimum height=8mm,
                align=center},
    circ/.style={draw, circle, minimum size=14mm},
    circInner/.style={draw, circle, minimum size=7mm},
    lbl/.style={font=\scriptsize, align=center},
    elbl/.style={font=\scriptsize, align=center}
]

\node[elbl] (cam) {EVENT\\CAMERA};

\node[box, right=of cam, font=\scriptsize] (sf) {SPATIAL\\FILTER};

\draw[->] (cam.east) -- node[midway, above, align=center, font=\scriptsize] {30K\\events} (sf.west);

\node[circ, right=of sf] (cb1) {};
\node[lbl] at (cb1.center) {CIRCULAR\\BUFFER};

\draw[->] (sf.east) -- node[midway, above, align=center, font=\scriptsize] {1K\\events} (cb1.west);

\node[box, right=of cb1, font=\scriptsize] (gaukf) {GA-UKF};

\node[circ, below=6mm of cb1] (cb2) {};
\node[lbl] at (cb2.center) {SHARED\\CIRCULAR\\BUFFER};

\node[box, left=of cb2, font=\scriptsize] (dec) {DECODER};
\node[elbl, left=of dec] (out) {FINAL\\OUTPUT\\TEXT};

\draw[->] (cam.east) -- (sf.west);
\draw[->] (sf.east) -- (cb1.west);
\draw[->] (cb1.east) -- (gaukf.west);

\draw[->] (gaukf.south) |- node[midway, left, align=center, font=\scriptsize] {ENCODED\\BITS} (cb2.east);

\draw[->] (cb2.west) -- (dec.east);
\draw[->] (dec.west) -- (out.east);

\end{tikzpicture}
}
\caption{\small Event pipeline: spatially filtered events are buffered, tracked with GA-UKF, and decoded via a shared buffer to reconstruct binary data in real time.}
\label{fig:system_architecture}
\end{figure}
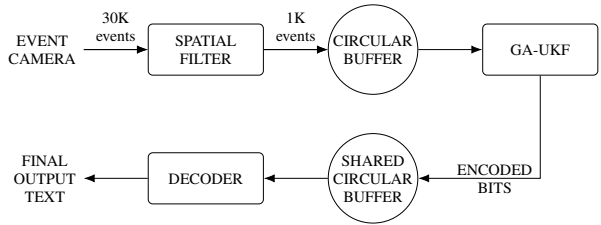

Our overall approach is shown in Fig.~\ref{fig:system_architecture}. Events from the camera are first preprocessed using a spatial filter, reducing the event rate from approximately 30k to 1--2k events per second. The spatial filter divides the image into fixed grid cells with precomputed cell assignments and neighbor relationships, and processes only locally active cells and their neighbors, helping remove background noise and focus on the relevant regions. This reduces computational load and keeps the batch updates stable for the subsequent state estimation. Filtered events are stored in a circular buffer accessed by the tracking node, which performs state estimation using GA-UKF to predict the LED's position and dynamically define the region of interest for the next step. Binary data extracted from the tracked ROI using~\cite{wang2022smart} is passed to the decoding node via a shared buffer, which reconstructs the transmitted message. This pipeline architecture enables concurrent tracking and decoding while maintaining real-time performance.



\subsection{Geometry-aware state estimation}
As mentioned earlier, the covariance matrix, which encodes the geometric structure of the blob, should be processed in a non-Euclidean space. This geometric formulation provides a principled process model without relying on numerical hard constraints to maintain a valid state (e.g., enforcing positivity of the semi-axis lengths). Since the covariance matrix lies on the manifold of symmetric positive definite (SPD) matrices, its evolution is naturally governed by the geodesic equation under the affine-invariant Riemannian metric\cite{bhatia2007positive}. We further introduce a product manifold approximation of the SPD manifold to reduce computational complexity and provide a simplified solution. 

\subsubsection{Geometry of Ellipse Parameters}
The blob covariance matrix $\Sigma \in \mathbb{S}_{++}^2$ can be parameterized as an ellipse via eigen-decomposition as follows:


\begin{equation}
    \begin{aligned}
        \Sigma = \mathbf{R}(\theta) 
        &\begin{bmatrix} 
        \lambda_1^2 & 0 \\ 
        0 & \lambda_2^2 
        \end{bmatrix}
        \mathbf{R}^\top(\theta), \\[6pt]
        \text{where} \quad
        \mathbf{R}(\theta) =
        &\begin{bmatrix} 
        \cos\theta & -\sin\theta \\ 
        \sin\theta & \cos\theta 
        \end{bmatrix}.
    \end{aligned}
    \label{eq:ellipse_param}
\end{equation}
Here $\lambda_1, \lambda_2 \in \mathbb{R}^+$ are the semi-axes and $\theta \in \mathbb{S}^1$ is the orientation angle. For a very small step, movement over the manifold can be considered geodesic.

The manifold $\mathbb{S}_{++}^2$ evolves according to geodesic flow. Thus, a small curve length $ds$ with respect to the affine-invariant Riemannian metric can be described as~\cite [Chapter~6]{bhatia2007positive}:
\begin{equation}
ds^2 = \mathrm{tr}(\Sigma^{-1} d\Sigma \, \Sigma^{-1} d\Sigma).
\label{eq:affine_metric}
\end{equation}
where $d\Sigma$ denotes the infinitesimal variation of the covariance matrix along a smooth curve on the $\mathbb{S}_{++}^2$ manifold, and $ds$ represents the infinitesimal geodesic step along this curve. For the ellipse parameterization~\eqref{eq:ellipse_param}, this metric yields:
\begin{equation}
ds^2 = 4\,d\mu_1^2 + 4\,d\mu_2^2 + 2\left(\frac{\lambda_1^2 - \lambda_2^2}{\lambda_1 \lambda_2}\right)^2 d\theta^2,
\label{eq:coupled_metric}
\end{equation}
where $\mu_i = \log \lambda_i$ are logarithmic coordinates for $i=1, 2$. The affine-invariant formulation requires repeated matrix logarithm/exponential operations and eigen-decompositions, leading to higher computational cost. We therefore adopt a product manifold representation which is simpler, computationally efficient, and achieves high accuracy in our application.

\subsubsection{Product Manifold Approximation}

Rather than operating on the full SPD manifold with the affine-invariant metric \eqref{eq:affine_metric}, we adopt a local approximation valid for small time steps. From~\eqref{eq:coupled_metric}, the logarithmic axis-length coordinates $\mu_i = \log \lambda_i$ have constant metric coefficients and evolve independently in log-space, while the orientation component $\theta$ is influenced by the ellipse eccentricity.
For a fixed-shape LED as in our application, eccentricity may vary due to events caused by high-velocity motion along particular directions. However, when processed over a small time interval, such variations do not lead to degeneration of eccentricity, and it remains bounded even at high velocities. Hence,in an infinitesimal time window ,influence of axis lengths on orientation change  can be modeled with stochastic noise, independent of semi-axis lengths. 

The local geometry can thus be approximated by a product manifold as
\begin{equation}
\mathbb{M} = \mathbb{R} \times \mathbb{R} \times \mathbb{S}^1,
\label{prod_manifold}
\end{equation}
with decoupled coordinates $(\mu_1, \mu_2, \theta)$ and simplified metric
\begin{equation}
ds_{\text{prod}}^2 = d\mu_1^2 + d\mu_2^2 + d\theta^2.
\end{equation}
This corresponds to a first-order tangent-space approximation of the affine-invariant metric. Under the standard tangent decomposition of product manifolds~\cite[Proposition~3.14]{lee2013smooth}, filtering operations can be performed independently in each factor space, reducing computational complexity while preserving local geometric consistency. 

\subsubsection{State Space Formulation}
Considering the geometric characteristics described above, we define our state vector as
\begin{equation}
    \mathbf{x} = [x, y, v_x, v_y, \lambda_1, \lambda_2, \theta, \omega]^\top,
    \label{eq:gaukf_state_vector}
\end{equation}
where $(x, y)$ and $(v_x, v_y)$ are the Euclidean position and velocity components, respectively. The semi-axis lengths $(\lambda_1, \lambda_2)$ and orientation $\theta$ evolve on the product manifold $\mathbb{R}^2 \times \mathbb{S}^1$, with the axis lengths evaluated in log space. The angular velocity $\omega$ exists in Euclidean space and captures changes in the blob's motion direction and ego-rotation. This separation of Euclidean and manifold components enables geometrically consistent GA-UKF filtering while accounting for both translational and rotational dynamics.


\subsubsection{Axis Length Prediction}
Semi-axis lengths of the blob encode information about its motion. The major axis length tends to increase along the direction of transmitter motion. For a fixed-shape blob observed over short event-time windows, changes in minor axis length primarily reflect changes in depth. Therefore, depth-induced variations in axis length can be disambiguated by examining recent changes in the minor axis length. We compute this change as
\begin{equation}
    \delta \lambda_{\mathrm{minor}}
    =
    \log \frac{\lambda_{k-1}}{\lambda_{k-2}},
    \label{eq:motion_inc}
\end{equation}
where $\lambda_{\mathrm{minor}}$ represents the minor-axis length expressed in log space, and $(k-1, k-2)$ denote the two previous states. Driven solely by relative axis-length changes, this effect is orientation-invariant. By incorporating it into the process model \eqref{proc_mod}, it improves axis length prediction under high measurement noise. A demonstration is provided in the supplementary video.

\subsection{GA-UKF Implementation}

Building on the product-manifold formulation from the previous section, we extend the standard Unscented Kalman Filter (UKF)~\cite{julier2004unscented}. Instead of mixing quantities across different manifolds during covariance computations, all manifold-valued components are locally mapped to their Euclidean tangent spaces via the product metric~\eqref{prod_manifold}. This ensures that sigma-point propagation, covariance, and cross-covariance calculations remain consistent while preserving the underlying manifold geometry. Algorithm~\ref{alg:geo_ukf_compact} presents the step-wise details of the proposed GA-UKF. 

\begin{algorithm}
\begin{algorithmic}[1]

\Require Current state mean and covariance
\Ensure Updated state mean and covariance

\State Run UKF \textbf{Prediction through Process Model as in~\ref{proc_mod}:}
\begin{itemize}[noitemsep]
    \item Axes lengths in log-space
    \item Orientation wrapped on $\mathbb{S}^1$, evolves with angular velocity 
\end{itemize}

\State Run UKF \textbf{Measurement:}
\begin{itemize}[noitemsep]
    \item Fuse temporally weighted as per eq.\ref{meas_mod}
    \item Axes lengths in log-space, orientation on $\mathbb{S}^1$
\end{itemize}

\State \textbf{Update} state belief:
\begin{itemize}[noitemsep]
    \item Euclidean components updated normally
    \item Axes lengths updated from log-space as per eq.~\ref{eq:lambda_update}
    \item Orientation wrapped on $\mathbb{S}^1$
\end{itemize}

\end{algorithmic}
\caption{GA-UKF for Elliptical Blob Tracking}
\label{alg:geo_ukf_compact}
\end{algorithm}

\subsubsection{Process Model for GA-UKF}
The state follows constant-velocity kinematics, i.e., the position evolves according to velocity while velocity remains unchanged. The blob shape and orientation evolve according to geometry-aware updates. Semi-axes are updated in the logarithmic domain by motion-induced changes ~\eqref{eq:motion_inc} with noise. Orientation evolves via angular velocity, accounting for ego-rotation and motion direction. All state elements evolve with Gaussian noise.
\begin{equation}
\mathbf{X}_{k+1|k}^{(i)} =
\begin{bmatrix}
x + v_x \Delta t \\ 
y + v_y \Delta t \\ 
v_x \\ 
v_y \\ 
\log\lambda_1 + \delta \lambda_{\text{minor}}\\ 
\log\lambda_2 + \delta \lambda_{\text{minor}} \\ 
\mathrm{atan2}(\sin(\theta + \omega \Delta t), \cos(\theta + \omega \Delta t)) \\ 
\omega
\end{bmatrix}^{(i)} 
+ \mathbf{q}_k,
\label{proc_mod}
\end{equation}
where $\mathbf{q}_k \sim \mathbb{N}(\mathbf{0}, \mathbf{Q})$ represents Gaussian process noise.

\subsubsection{Temporally Weighted Measurements for GA-UKF}
A blob formed by $N$ events from a blinking transmitter is spatially approximated as a 2D Gaussian. Each event is assigned a temporal weight
\[
w_j = \exp(-\beta \, (t_k - t_j)), \quad j=1,\dots,N,
\] 
giving higher importance to recent events. This improves the measurement of the blob shape and preserves it when the buffer includes older motion-induced events. The weighted mean and covariance are
\begin{equation}
\mu_z = \sum_{j=1}^N w_j 
\begin{bmatrix} x_j \\ y_j \end{bmatrix}, \quad
C_e = \sum_{j=1}^N w_j (\mathbf{r}_j - \mu_z)(\mathbf{r}_j - \mu_z)^\top,
\label{weights}
\end{equation}
where
\[
\mathbf{r}_j = 
\begin{bmatrix} x_j \\ y_j \end{bmatrix}, \quad j=1, \ldots, N events,
\]
and $x_j, y_j$ are pixel coordinates. Semi-axis lengths $\lambda_1, \lambda_2$ and orientation $\theta$ are derived from $C_e$, giving the measurement vector
\begin{equation}
z_k =
\begin{bmatrix}
x_{\text{blob}} & y_{\text{blob}} & \lambda_1 & \lambda_2 & \theta
\end{bmatrix}^\top \in \mathbb{R}^5.
\label{meas_mod}
\end{equation}
\subsubsection{Semi-Axis Length Update}
The semi-axes are updated in log-space to ensure positivity and numerical stability. For each axis $j=1,2$, the update is computed as


\begin{equation}
\lambda_j^{(k)} 
= \exp \Big(
    \log \lambda_j^{(k|k-1)} 
    + \mathbf{K}_{\lambda_j} 
      \big( \mathbf{z}_k - \hat{\mathbf{z}}_k \big)
\Big),
\label{eq:lambda_update}
\end{equation}

\noindent
where $\lambda_j^{(k|k-1)}$ is the predicted semi-axis at time $k$ (prior),
$\lambda_j^{(k)}$ is the updated semi-axis (posterior),
$\mathbf{K}_{\lambda_j}$ is the row of the Kalman gain corresponding to $\lambda_j$,
$\mathbf{z}_k$ is the measurement at time $k$, and
$\hat{\mathbf{z}}_k$ is the predicted measurement.
\subsection{Decoding Optical Signals}

For decoding optical signals, we employ the asynchronous event-based demodulation approach proposed by Wang et al.~\cite{wang2022smart}. In this method, the asynchronous event stream is converted into a continuous-time signal as given in equation (1) of ~\cite{wang2022smart}.

Fig.~\ref{fig:finalIllus} illustrates the decoding mechanism under motion conditions. The left panel depicts two consecutive time instances where GA-UKF continuously estimates the LED position and shape. The green dashed circles represent all events generated within the spatial neighborhood, the red circles indicate the actual LED state at any given time, and the purple dashed circles show the estimated blob state through GA-UKF. During motion, the total event stream (green) comprises contributions from three sources: LED intensity modulation (the communication signal), motion-induced brightness changes as the LED traverses the image plane, and ambient background noise. Consequently, the spatial event density exceeds what would be observed from a stationary flickering LED.



The right panel of Fig.~\ref{fig:finalIllus} illustrates the resulting continuous-time signal, where the horizontal lines represent the upper and lower decision thresholds. Binary signal reconstruction is achieved using a dual-threshold decision rule with hysteresis. This decoding process robustly recovers the transmitted binary sequence despite motion-induced artifacts. The reconstructed continuous signal $s(t)$ integrates event polarities with temporal decay, producing oscillations corresponding to the LED's ON/OFF modulation states. Critically, the shaded buffer region between upper and lower threshold provides immunity against spurious bit transitions. Motion-induced events and background noise modulate the signal amplitude but remain insufficient to trigger false threshold crossings, as the hysteresis mechanism requires sustained polarity accumulation to register state changes. This spatial-temporal filtering synergy, which combines the GA-UKF's state estimation with the decode, enables reliable communication even under significant relative motion.

\section{Experimental Results}
We present experimental results evaluating real-time optical communication performance on physical hardware. The experiments assess the impact of key system parameters, including signal frequency, camera to LED distance  and relative motion, on communication accuracy. 

The primary metrics used for evaluation are:
\begin{inparaenum}[i)]
    \item \emph{Accuracy:} defined as the ratio of correctly decoded words to the total number of transmitted words;
    \item \emph{Processing time per data packet:} measures the computational efficiency of the system; and
    \item \emph{Robustness to motion:} evaluates the system's ability to handle varying motion dynamics without degradation in performance.
\end{inparaenum}

\subsection{Hardware and Software Setup}
Experiments were conducted on an Intel i5-13600 (x86\_64) processor and a Jetson Orin NX 16GB using a Prophesee EVK4 event camera with an 8\,mm lens and an aperture ranging from f/2.8 to f/8. Standard LEDs driven by an ESP32 generated blinking signals, with transmitters mounted on the drone~\ref{fig:hardware}. For higher brightness, a SparkFun LP55231 LED driver was used; however, its 1\,kHz frequency limit required ESP32-controlled LEDs for high-frequency tests. Event data was processed in 4\,ms windows. At such short intervals and moderate velocities, temporally weighted measurements have minimal effect. However, as per~\eqref{weights}, larger weights become important when using longer time windows or operating at high velocities. They help in preserving the blob shape by emphasizing recent events. 


\subsection{Experiments}
To test the capabilities of our communication pipeline, we pair a stationary event camera with a moving frequency-modulated LED transmitter in three hardware configurations: (i) LED mounted on a drone following a random trajectory, (ii) rapid handheld LED motion in indoor and outdoor environments, and (iii) a circular disk rotating at high angular velocity with the LED mounted on the edge. We further tested the system on a Jetson Orin NX 16 GB to assess embedded performance and ego-motion handling (see supplementary video).

\begin{figure} [h]
    \centering
    \includegraphics[width=1\linewidth]{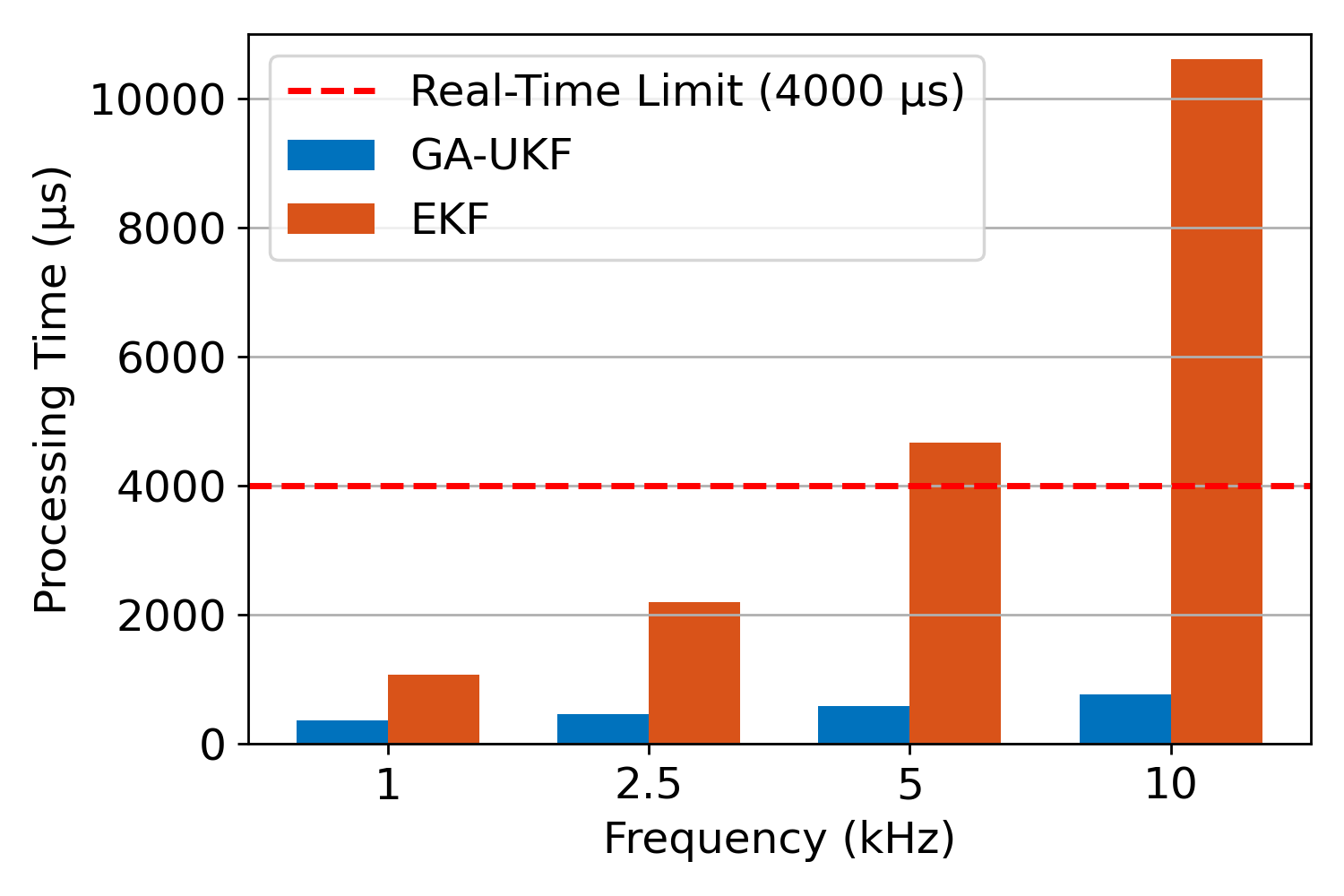}
    \caption{Processing time comparison between GA-UKF and EKF from \cite{async_blob_wang} at varying signal frequencies.}
    \label{fig:proctimes}
\end{figure}

\begin{figure*}
\centering

\begin{mdframed}\begin{lstlisting}
The farthest known star ever observed from Earth is called Earendel, and the light we see from it today began its journey nearly thirteen billion years ago, meaning we are observing the star as it existed when the universe was less than one billion years old. Earendel is visible only because of gravitational lensing, where a massive galaxy cluster bends and magnifies its light, acting like a natural cosmic telescope that makes an otherwise invisible star detectable across vast distances. Due to the expansion of the universe, the star's present-day distance is estimated to be around twenty-eight billion light-years.\end{lstlisting}\end{mdframed}

\begin{mdframed}\begin{lstlisting}
The farthest know|r|'>o'\{?\&o+'J\^|r| Earth is called Earendel, and the light we see from it |r|todegan|r| its journey nearly thirteen billion years ago, meaning we are observing the star as it existed when the universe was less than one billion years old. Earendel is visible only because of gravitational lensing, where a massive galaxy cluster bends and magnifies its light, acting like a natural cosmic telescope that makes an otherwise invisible star detect1ble across vast distances. Due to the expansion of the universe, the stars present-day distance is estimated to be around twenty-eight billion light-years.\end{lstlisting}\end{mdframed}

\begin{mdframed}\begin{lstlisting}
|r|`\$10 l\$!0! 1'```10 f! ` `s`toH` tn0(t`ol(t6't`9wo0w6v`` `1'0k o\#`la<>C` `30`'=l\$(w~tE\_\$w+`o`>w6`\$m\{`3                                                                                                     9f`n`0sEt1ts``clte`30tC1`oXelCEC?og`wEll\&gn`np[                                                                                                                                                    +_>`.lo`yl,`lo\{C``` x`w`'\$`t`'+w~n          9Y]gi>o0olld8s"|`w?0(~``|6<|r|\end{lstlisting}\end{mdframed}

\caption{Original text (top), compared to decoded text using GA-UKF (middle) and EKF (bottom). Errors highlighted in red.}\label{fig:textcomparison}
\end{figure*}

Figure~\ref{fig:proctimes} demonstrates the computational efficiency of GA-UKF compared to the baseline EKF \cite{async_blob_wang} across varying signal frequencies. Using a 4~ms time window for event packets, GA-UKF maintains processing times consistently below the 4000~$\mu$s real-time threshold. Exceeding this threshold would cause new data to enter the buffer before the previous packet is processed, breaking continuity and corrupting the buffers (see Fig.~\ref{fig:system_architecture}). GA-UKF achieves a maximum of 763~$\mu$s at 10~kHz, whereas EKF exceeds the threshold above 4~kHz, reaching 10,603~$\mu$s at 10~kHz—over 13$\times$ slower. Figure~\ref{fig:textcomparison} highlights the impact on decoding accuracy: GA-UKF reconstructs the reference text with few character errors, while EKF produces garbled output. These results confirm that GA-UKF enables reliable high-frequency optical communication under real-time constraints, whereas EKF fails under similar conditions.

\begin{figure}[h]
    \centering
    \begin{tikzpicture}
        \begin{axis}[
            name=left,
            width=0.75\linewidth,
            height=7cm,
            xlabel={Range (m)},
            ylabel={Accuracy (\%)},
            xmin=2, xmax=20,
            ymin=0, ymax=100,
            ytick={0,25,50,75,100},
            xtick={3, 6, 9, 12, 15, 18},
            xticklabel style={/pgf/number format/1000 sep={}},
            grid=both,
            tick align=outside,
            line width=1pt,
            legend style={
                at={(0.02,0.02)},
                anchor=south west,
                font=\scriptsize,
                draw=none
            },
            axis lines=box,
            after end axis/.code={
            \draw[white, line width=2pt]
                    (rel axis cs:1,0) -- (rel axis cs:1,1);
            \begin{scope}[xshift=4.5pt] 
                \draw[white, line width=4pt]
                    ([xshift=0pt, yshift=-3pt]rel axis cs:1,1) --
                    ([xshift=0pt, yshift=3pt]rel axis cs:1,1);
                \draw[black, line width=1pt]
                    ([xshift=-3pt, yshift=-3pt]rel axis cs:1,1) --
                    ([xshift=3pt,  yshift=3pt]rel axis cs:1,1);
                \draw[black, line width=1pt]
                    ([xshift=-1pt, yshift=-3pt]rel axis cs:1,1) --
                    ([xshift=5pt,  yshift=3pt]rel axis cs:1,1);
                \draw[white, line width=4pt]
                    ([xshift=0pt, yshift=-3pt]rel axis cs:1,0) --
                    ([xshift=0pt, yshift=3pt]rel axis cs:1,0);
                \draw[black, line width=1pt]
                    ([xshift=-3pt, yshift=-3pt]rel axis cs:1,0) --
                    ([xshift=3pt,  yshift=3pt]rel axis cs:1,0);
                \draw[black, line width=1pt]
                    ([xshift=-1pt, yshift=-3pt]rel axis cs:1,0) --
                    ([xshift=5pt,  yshift=3pt]rel axis cs:1,0);
            \end{scope}
            },
        ]

        \addplot[gray, dashed] coordinates {(2,90) (20,90)};
        \addlegendentry{90\% accuracy}

        \addplot[red, mark=square*, thick]
        coordinates {(2,92) (5,93.5) (8,94) (12,95) (18,93.5)};
        \addlegendentry{Bright LED @ 1kHz}

        \addplot[blue, mark=*]
        coordinates {(2,93) (5,94) (8,95) (12,94.5) (18,60)};
        \addlegendentry{Normal LED @ 1kHz}

        \addplot[black, mark=diamond*]
        coordinates {(2,94) (5,92) (8,95) (12,95.5) (18,65)};
        \addlegendentry{Normal LED @ 2.5kHz}

        \addplot[purple, mark=triangle*]
        coordinates {(2,95) (5,93) (8,91) (12,90) (18,67)};
        \addlegendentry{Normal LED @ 5kHz}

        \addplot[yellow!80!black, mark=o]
        coordinates {(2,94.5) (5,92) (8,93) (12,89) (18,63)};
        \addlegendentry{Normal LED @ 10kHz}

        \end{axis}

        \begin{axis}[
            name=right,
            at={($(left.south east)+(0.4cm,0)$)},
            anchor=south west,
            width=0.3\linewidth,
            height=7cm,
            xlabel={},
            ylabel={},
            yticklabels={},
            xmin=55, xmax=65,
            ymin=0, ymax=100,
            ytick={0,25,50,75,100},
            xtick={60},
            grid=both,
            tick align=outside,
            line width=1pt,
            axis lines=box,
            after end axis/.append code={
              \draw[white, line width=2pt]
                (rel axis cs:0,0) -- (rel axis cs:0,1);
            },
        ]

        \addplot[gray, dashed] coordinates {(55,90) (65,90)};

        \addplot[red, mark=square*, thick]
        coordinates {(60,91)};

        \addplot[blue, mark=*]
        coordinates {(60,2)};

        \addplot[black, mark=diamond*]
        coordinates {(60,2)};

        \addplot[purple, mark=triangle*]
        coordinates {(60,2)};

        \addplot[yellow!80!black, mark=o]
        coordinates {(60,2)};

        \end{axis}
    \end{tikzpicture}
    \caption{Accuracy vs Distance between Camera and LED for different LED configurations and frequencies.}
    \label{fig:accvsdist}
\end{figure}
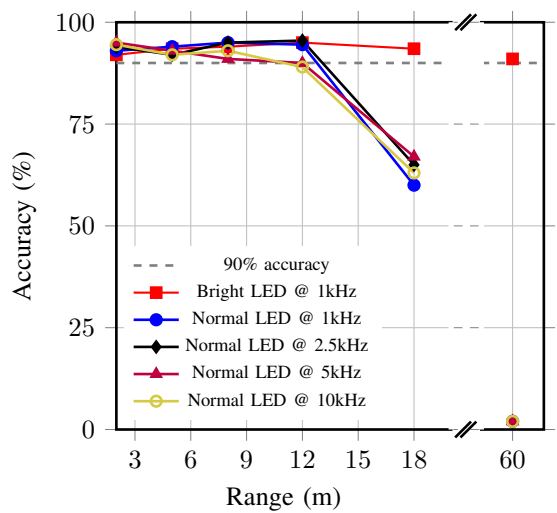

Figure ~\ref{fig:accvsdist} presents the relationship between communication accuracy and transmission range for different LED configurations under dynamic conditions. Due to hardware limitations, bright LED experiments were restricted to lower frequencies (1 kHz). Bright LEDs help maintain high accuracy beyond 60\,m, whereas normal LEDs degrade the accuracy rapidly past 15\,m due to limited intensity. Thus, while our system demonstrates robust performance under dynamic conditions, long-range communication remains hardware-limited and requires higher light output to compensate for distance-related attenuation.

This system targets real-world scenarios, which typically exhibit lower pixel velocities (e.g., a drone at 20~m/s and 8~m produces $\approx$ 4100~px/s). To stress-test at higher speeds, we mounted the LED on a circular disk with a motion path diameter of $\approx$ 610~pixels, generating faster relative motion. GA-UKF maintains over 90\% accuracy up to $\approx$ 5,000~px/s (Table~\ref{tab:speedtable}). Between 6,000 – 7,500~px/s, only minor errors occur (typically 1–2 missing characters per word), so the words remain mostly readable. Beyond 10,000~px/s, accuracy drops sharply due to motion blur, which produces spurious events. Since our metric evaluates complete words, a single incorrect bit renders an entire word wrong. Up to 8,000~px/s, the character-level transducer accuracy still stays above 90\%. Our experiments also showed that higher signal frequencies slightly improve accuracy as motion-blur event volume is smaller in comparison to signal event volume in this case (see table~\ref{tab:speedtable}).

 \subsection{Scope of Improvement}
Although the proposed framework remains robust at high velocities, we observe several failure modes: (i) sudden jerks or rapid accelerations can amplify estimation errors, particularly at high speeds; (ii) bright background lights may induce short-term decoding errors; and (iii) overlapping transmitters can generate incorrect messages and temporarily inflate the covariance as they move away from each other, potentially leading to loss of tracking.

While the current setup can handle high-velocity target motion, camera motion must remain within the transmitter frame for accurate tracking. Tracking performance can be further improved, particularly in scenarios with significant ego-motion and rotation, by incorporating IMU measurements, which can aid in self-prediction and reduce errors caused by jerky motion.
\begin{table}
\centering
\footnotesize
\begin{tabular}{|c|c|c|c|}
\hline
\multirow{2}{*}{\centering\textbf{Speed (pixel/s)}} & \multicolumn{3}{c|}{\textbf{Accuracy}} \\
\cline{2-4}
 & \textbf{Freq. 1 kHz} & \textbf{Freq. 5 kHz} & \textbf{Freq. 10 kHz} \\
\hline
1500 & 94\% & 96\% & 97\% \\
\hline
3000 & 92\% & 95\% & 96\% \\
\hline
4500 & 91\% & 95\% & 95\% \\
\hline
6000 & 73\% & 82\% & 86\% \\
\hline
$>10000$ & $<$10\% & $<$10\% & $<$15\% \\
\hline
\end{tabular}
\caption{Decoding accuracy with GA-UKF at different speeds and frequencies}
\label{tab:speedtable}
\end{table}

\section{Conclusion}
We demonstrated a complete optical communication system using event cameras and high-frequency LED signaling. Our Geometry-Aware UKF algorithm enables real-time processing by batching events and exploiting the structured geometry of blob tracking. The 7× speedup over EKF~\cite{async_blob_wang} makes high-frequency optical communication practical for mobile robotics applications. We built our approach on strong geometric foundations, combined with statistical-geometric connections established in multivariate analysis~\cite{johnson2007applied} to enable efficient implementation without sacrificing accuracy.

In future, our proposed framework can be extended to broader multi-agent scenarios, including simultaneous multi-target tracking, visual–inertial fusion for robustness under dynamic camera motion, decoding under varying transmitter frequencies, and integration with control and navigation objectives such as formation control.
\section{Supplementary Video}
Supplementary video for this work is available at: \\
\url{https://drive.google.com/file/d/139rS0r0Fm2KQv4UwHfqsGmnNULe-qH0h}

\bibliographystyle{IEEEtran}
\bibliography{references}

\end{document}